\documentclass[10pt, a4paper]{article}
\usepackage{lrec2022} 
\usepackage{multibib}
\newcites{languageresource}{Language Resources}
\usepackage{graphicx}
\usepackage{tabularx}
\usepackage{soul}
\usepackage{titlesec}
\titleformat{\section}{\normalfont\large\bfseries\center}{\thesection.}{1em}{}
\titleformat{\subsection}{\normalfont\SmallTitleFont\bfseries\raggedright}{\thesubsection.}{1em}{}
\titleformat{\subsubsection}{\normalfont\normalsize\bfseries\raggedright}{\thesubsubsection.}{1em}{}
\renewcommand\thesection{\arabic{section}}
\renewcommand\thesubsection{\thesection.\arabic{subsection}}
\renewcommand\thesubsubsection{\thesubsection.\arabic{subsubsection}}

\usepackage{epstopdf}
\usepackage[utf8]{inputenc}

\usepackage{hyperref}
\usepackage{xstring}
\usepackage{mathtools}
\usepackage{color}

\usepackage{array, makecell, multirow, tabularx}
\newcolumntype{C}{>{\raggedright\arraybackslash}X}

\usepackage{enumitem}

\usepackage[normalem]{ulem}
\newcommand{\shwTrn}[1]{}

\newcommand{\citet}[1]{\newcite{#1}}
\newcommand{\citep}[1]{\cite{#1}}
\newcommand\newterm[1]{\textbf{#1}}
\newcommand\term[1]{\textit{#1}}

\newcommand{\vi}{\texttt{v1}} 
\newcommand{\vii}{\texttt{v2}} 

\newcommand{\pmx}{$\pm$}

\title{Thematic Fit Bits: Annotation Quality and Quantity Interplay \\for Event Participant Representation}
\name{Yuval Marton$^1$, Asad Sayeed$^2$}
\address{
  $^1$University of Washington, WA, USA 
  \\
  $^2$Dept. of Philosophy, Linguistics, and Theory of Science, 
  University of Gothenburg,
  Sweden 
  \\
  {ymarton@uw.edu, asad.sayeed@gu.se}
  \\
}

\abstract{
Modeling 
 thematic fit
(a verb--argument compositional semantics task) currently requires a very large burden of labeled data.  We take a
 linguistically machine-annotated large corpus and replace
corpus layers
with
output from higher-quality, more modern taggers.
We compare the old and new corpus versions' impact on a
verb--argument fit modeling task, using a
high-performing
neural approach.  
We discover that 
higher 
annotation quality dramatically reduces our data requirement while demonstrating better supervised predicate-argument classification. But in 
 applying the model to psycholinguistic tasks outside the training objective, we see clear gains at scale, but only in one of two thematic fit estimation tasks, and no clear gains on the other.
 We also see that quality improves with training size, but perhaps plateauing or even declining in one task.
 Last, we tested the effect of role set size.
All this suggests that the quality/quantity interplay is not all you need.
We replicate previous studies while modifying certain role representation details and set a new state-of-the-art in event modeling, using a fraction of the data. 
We make the new corpus version public.
 \\ \newline 
}

\begin{document}

\maketitleabstract

\section{Introduction}\label{sec:intro}
Is more data always more effective than better annotation?  Is it always cheaper just to obtain and use data with mid-quality annotation than improve annotation quality over a smaller dataset? 
Traditionally, to researchers grounded in linguistics, it seemed obvious that higher quality and richer annotation should be better. But with the advent of ``Big Data'', 
the common wisdom
seem to have shifted toward more data;
Deep Learning continued this trend (see examples in Section~\ref{sec:intro-data}).

We re-examine these questions
using two types of natural language processing (NLP) tasks:
\textbf{(1)} 
 supervised 
thematic role prediction
(given a predicate and an argument's word span; here only its syntactic head word)
and word prediction (given a predicate and a role); and
\textbf{(2)}
psycholinguistic tasks outside the explicit  training objective: rating the thematic fit between a verb and its potential arguments.
These tasks have a large body of work in computational linguistics (see section \ref{sec:semmodel}).

We examine the trade-offs  in training models 
designed to accomplish these tasks 
through modeling events and their participants in  a large corpus (task (1) above). The trade-off we focus on is using more data with 
mediocre
linguistic annotations versus little 
data with higher-quality 
annotations.
For the former, 
we replicate a PropBank-based model of \newcite{hong-2018-learn-event-rep} using increasing subsets of their training data,
a large corpus with machine-predicted annotations of mediocre quality. 
For the latter, we replace some 
annotation layers with  equivalent layers generated by higher quality linguistic tools.
Our model implementation differs from  \newcite{hong-2018-learn-event-rep} in 
how missing role information and unknown (out-of-vocabulary) roles are represented.
Our replicated baseline is stronger than theirs.

We also test whether  training on the higher quality data keeps yielding better models as training set size increases.
Last, we also look at the trade-offs in training models over increasingly richer, more fine-grained, but potentially sparser semantic role annotation.

\subsection{The conundrum of data}\label{sec:intro-data}

Our
first goal  
is to revisit a widely held assumption in the NLP community: mediocre machine-predicted annotation  yields at scale better (or equivalent) models than high quality annotations (manual or state-of-the-art machine-predicted) whose scale is much smaller, due to compute, cost, and time constraints. 
\newcite{mcclosky:2006-self-train} and \newcite{foster++:2007-BNC-self-train}, \textit{inter alia},
use self-training%
\footnote{Augmenting 
few
manual annotations with ones predicted by a prior version of the same parser over a large text.}
to improve syntactic parsing -- as an alternative to manually annotating more data -- in same or different domain/genre.
\newcite{petrov++:2010:uptraining} show that using  100k machine-predicted constituency parses to train a new dependency parser contributed the equivalent of 
2k manually annotated parses. Manually annotating is slower and more expensive but better by definition.

However, despite growing amounts of annotated and unannotated textual resources, a number of tasks with traditional linguistic levels of representation remain a challenge, with unsatisfactory performance in various research areas and applications: artificial intelligence (AI), machine reading / knowledge graph population, chatbots and natural language understanding (NLU), as well as computational psycholinguistic modeling and computational linguistics.

How do quality and quantity affect our above-mentioned semantic and psycholinguistic tasks?

\subsection{Semantic modeling}
\label{sec:semmodel}
Our second goal 
is to explore ways to improve semantic modeling and the representations used for this modeling.  We use \term{semantic modeling} to refer to 
tasks at the intersection of NLP and psycholinguistics that have to do with representing and processing generalized event knowledge \citep{pustejovsky1991,zarcone2011}
. The underlying tasks include:

\newterm{Semantic role labeling (SRL)}
is the task of annotating text according to semantic frames and their roles as defined in frameworks such as FrameNet, VerbNet, or PropBank~\cite{framenet-Baker++:1998,verbnet-SchulerPalmer:2005,propbank-PalmerGildeaKingsbury:2005}.
For example, given `I cut the cake with...' (1) `marzipan' or (2) `a knife',  the role
\term{Instrument/A3-MNR} is normally desired for (2) but not for (1).

\newterm{Role prediction}:
a simplified SRL task we use here. Instead of a sentence, the input is just a verb $v$ and a noun $n$; optionally additional nouns and their thematic relation (role) with $v$; the expected output is the most likely thematic role of $n$ with $v$ in the same event.%
\footnote{A similar task -- only without the input nouns -- is the
    \newterm{(role) selectional preference} of the verb: what roles are more likely with this verb? E.g., `cut': \term{Agent/Arg0, Patient/Arg1, Instrument/Arg3-MNR}.
}

\newterm{Slot / Role filling}:%
\footnote{We interchangeably refer to it here as \newterm{word prediction}.} 
given a predicate (typically a verb) and a thematic role, what word or phrase would be most appropriate for that role? This task can be viewed as the complement of SRL (given the word or phrase, what is its role?).

\newterm{Thematic fit}:
Given a predicate and a role (say, `cut' + \term{Instrument}), how well would a speaker of a given language (here, English) find `knife' or `spoon' 
fitting to the given role?
And by extension:
given a 
subset of a predicate and arguments (optionally also modifiers), can we predict the typicality level of the most recently added member to the rest of the given subset? This is often an abstraction of sentence comprehension (in humans): our {\it thematic fit} estimation changes as we hear more of the uttered sentence \citep{amsel2015close}. 

Indirect thematic fit estimation learning from SRL annotations has shown promising results \cite{hong-2018-learn-event-rep,tilk2016event,santus2017measuring}. 
Following \newcite{hong-2018-learn-event-rep} and others, we consider only the arguments' syntactic heads together with their semantic roles.%
\footnote{We also follow them in using the simplified ProbBank roles (\term{A0, A1, ...}), unlike much of the thematic fit literature that uses \term{Agent, Patient,} etc. \cite{dowty1991thematic}.
}
Given a new role, 
predict the fitness level of all known words and 
use the score of the given filler in the full score distribution as a fitness rating. (We also look at the complement: given a new word, predict the fitness of each possible role). These predictions are scored relative to human 
judgments (see Section~\ref{sec:bg}).

\subsection{Contributions}
Exploring the data requirements of modeling human semantic representations allows us to revisit the question of the inherent difficulty of semantic tasks: 
does semantic processing simply require more annotated data to achieve high quality, or 
has it not been represented in a way conducive for computers to learn adequately? 

We look into how annotation quality and 
quantity (both the number of semantic frames 
and the number of sentences) 
affect learning. We also explore annotation granularity,
taking thematic role set granularity (number of roles in model) as a test case.
We often see that modelers  focus on only the two most frequent PropBank 
semantic arguments (Arg0 and Arg1) and ignore the rest or lump the rest together under a catch-all tag. Similarly, they focus on only few modifiers (e.g.,
\newcite{tilk2016event} and \newcite{hong-2018-learn-event-rep} use 2 core roles and 3 modifiers). 
We therefore trained models with increasing numbers of 
thematic role types
 (the predicate, its core arguments, and often-optional, non-core arguments or modifiers), using the better taggers, parsers, and labelers, and observed changes in prediction quality.
 
To summarize our main contributions, we
\vspace{-9pt}
\begin{enumerate}[noitemsep]
    \item test how quality of annotation affects supervised role/word prediction, as training set size increases. We show in small to large sizes that the mediocre  annotation method 
    is \textit{not} as useful as better quality annotation.
     
    \item test how quality of annotation affects thematic fit estimation as an application of our models that is not part of the training objective. 
    We show  that quality increases with training size but surprisingly, variance is high even in larger sizes, leaving no clear winner annotation.

    \item claim that the high variance of the (indirectly optimized for) thematic fit estimation makes it more difficult to interpret conclusions from previous studies that did not report it. 
    
    \item show new state-of-the-art results on role and word prediction, as well as thematic fit estimate correlations.

    \item tease apart effects of quality and quantity;  
    tease apart 
    the number of training sentences from the number of training frames (the semantic frames annotated in these sentences).

    \item  test how annotation  granularity (role set size)
    affects thematic fit (and role/word prediction)

    \item provide a new, open-data, large lexical semantic (and syntactic) resource in English, revising and expanding the previously published RW-Eng \cite{sayeed2018}.

\end{enumerate}

\section{Background and related work}\label{sec:bg}
\paragraph{Thematic fit norms}
take the form of averaged human-rated plausibility scores for a verb, a noun, and the noun's thematic role.
For example, we ask human raters: how well does "sword" fit as an instrument with
the verb "cut"?  Thematic fit norms are a subset of semantic feature/property fit norms that pertain to verb-argument relations.  Exploring thematic fit allows for exploring the structure of the human lexicon and for exploring generalizations about affordances and the relationship between world knowledge and compositional semantics.

\citet{mcrae1998modeling} collected an early set of thematic fit norms. 
Human raters were asked 
to use a 7-point Likert scale
to judge the fit of particular nouns with particular verbs in given roles.  These  plausibility judgements focused mainly on \term{Agent-Patient} roles.  Later \citet{ferretti2001integrating} provided norms for \term{Instrument} and \term{Location} roles.  

\citet{pado2007integration} and \citet{pado2009probabilistic} sought to develop a probabilistic model of thematic fit.  In the process, they collected additional \term{Agent-Patient} norms for a limited, balanced subset of verb-noun pairs chosen by frequency in the Penn Treebank \cite{PTB-Marcus++:1993}.    Together, in addition to later efforts focusing on verb polysemy \citep{greenberg2015verb}, these collected norms form an empirical basis for modeling 
human semantic expectations, albeit limited to roles that are relatively frequent and easily understood by raters.

\paragraph{Distributional modeling of thematic fit}
Early work in thematic fit modeling  emphasized building partially or fully supervised corpus-based models \cite{pado2007dependency,herdagdelen2009bagpack}. The question arises whether less task-specific, less supervised models can be used to model the semantic generalizations that would underpin a robust thematic fit model. \citet{baroni2010distributional} proposed the Distributional Memory (DM) approach,  a very high-dimensional  tensor space representation that memorizes the frequency of numerous syntactic relations between lexical items in a large corpus  consisting of UkWaC \cite{ferraresi2008introducing}, the British National Corpus \cite[BNC]{british2007british}, and Wikipedia.  

\citet{sayeed2016thematic} applied the DM approach to relation features based in an early form of neural SRL tagger \cite[SENNA]{collobert2011natural}.  This and Baroni and Lenci's syntax-based features were combined synergistically to produce thematic fit correlation scores superior to the result of each individually. The DM models were constructed without any reference to the evaluation data and can be considered unsupervised in that sense.  However, their reliance on matrix multiplications made them difficult to extend to evaluating multiple roles simultaneously due to sparsity. They are also difficult to parameterize for finding optimal models.  

\citet{tilk2016event} and \citet{hong-2018-learn-event-rep} worked to supplant DM approaches with neural networks.  Their models train ``event'' embeddings with a preselected roleset, representing an entire semantic frame as input.  
Some
of the role ``slots'' can be left empty, allowing for a variable number of arguments to be tested.  \citet{hong-2018-learn-event-rep} applied a two-task training objective (limited SRL and role-filler noun prediction) to train NN models that not only performed well on thematic fit ratings, but also on several additional semantic tasks (e.g., event similarity and multiple-role compositionality).

\citet{sayeed2016thematic}, \citet{tilk2016event}, and \citet{hong-2018-learn-event-rep} all depend on the 
"Rollenwechsel-English", aka "RW-eng" corpus, hereafter \vi~\citep{sayeed2018}, 
and use almost all of the corpus to train their models. 
Our work builds on the work of \citet{hong-2018-learn-event-rep},
but differs in role set implementation,  some hyper-parameter settings, and minor other technical details.
Our work also builds on \vi\ and extends it with newer annotation layers.
One of the things we test is  whether the quantity of data used for these models is necessary to achieve those results, particularly on the 
thematic fit task.

\section{Dataset}\label{sec:dataset}
In order to explore the topics raised in Section~\ref{sec:intro}, we used the above-mentioned large-scale \vi\ corpus. 
It is annotated with a fast-but-outdated SRL tagger and syntactic parser. We added new annotation layers with higher quality, more modern taggers and parser (hereafter \vii, "Our annotations").
We replicated baseline models on \vi\  and trained new models on \vii, as detailed in Section~\ref{sec:experiments}.

\subsection{Text and \vi\  Annotations}
The \vi\ corpus 
\cite{sayeed2018}
consists of the SENNA-derived SRL output over 78M sentences from 2.3M documents.  The documents come from the BNC and ukWaC.  SENNA extracts multiple predicates per sentence and, for each predicate, it identifies spans of text representing noun phrases that fill PropBank 
roles for that predicate.  For every document, sentence, and predicate in that sentence, \vi\ 
contains XML-formatted information on the corresponding SENNA output.  In particular, it uses a series of head-finding heuristics \cite{sayeed2016thematic} 
to identify the syntactic heads of the role-filling spans---typically noun phrases, which can contain complex constituents such as subordinate clauses, but the SRL role spans could also cover only fractions of syntactic constituents, hence the need for a heuristic beyond only using a parser.

\subsection{Our Annotations (\vii)}
NLP often contains ``pipelines'' of serial annotation processes, such as: 
tokenization and morphological analysis (including lemmatization or stemming), 
syntactic parsing,
and a final processing such as machine translation, 
NLU (for chatbots), and sometimes SRL.
As mentioned in Section~\ref{sec:intro}, 
until about ten years ago, a rule of thumb often held: better quality of intermediate processing results in better quality at the end, although improvements are not linear, and small intermediate gains do not always translate to gains at the end of the pipeline.

Here we set to test out the new rule-of-thumb of the last decade for the case of thematic role prediction, slot filling (word prediction), and thematic fit estimation. We replaced annotation layers from older tools 
with annotations based on more recent tools (see below),
introducing a
non-negligible improvement in intermediate annotation quality.

The first step in doing so is to determine a consistent tokenization schema across all annotation layers, as some taggers either expect a certain schema or apply their own even if input text is largely tokenized.
We iteratively modified our tokenization schema to reduce token count mismatch between the new layers from over 20\% of sentences to less than 1\%. 
Once we reached this low mismatch ratio, we marked and excluded the fewer mismatched cases from our experiments.

We  added the following new annotation layers:

\paragraph{Lemmas by a newer morphological analyser: Morfette v0.4.4} \cite{chrupala-etal-2008-learning-morfette,chrupala-2011-efficient-morfette},
precompiled 
by 
Djam\'e 
Seddah,\footnote{We thank Djam\'e Seddah for making it available to us.} who included a transformed xtag lexicon \cite{seddah-spmrl-2013-overview}. Training was done on the Penn Treebank~\cite{PTB-Marcus++:1993}, using the Collins split.

\paragraph{Syntactic parses by a newer parser: spaCy 2.0.13} \cite{spacy-honnibal-johnson:2015:EMNLP,spacy2}.\footnote{https://spacy.io}
We forced spaCy to use our own tokenization 
instead of its own.

\paragraph{Semantic frames by a newer SRL tagger: LSGN} \cite{lsgn-he-2016-2018},  an end-to-end BiLSTM-based SRL tagger using ElMo embeddings \citep{Peters:2018}. It gets 86\% F1 score on the CoNLL05 WSJ test set, compared to SENNA's 75\%. 
For each semantic frame, we aligned the spaCy parses to each argument span in order to find the syntactic head of the span, using a similar heuristic as
in \vi.
We only used the heads for modeling (see Section~\ref{sec:experiments}).
We aligned each token in the argument span across all layers (surface word-form, Morfette lemma, spaCy lemma and entity (NER) tag, etc.)

\subsection{Train / dev / test split}\label{sec:split}
Of about 3500 files, few (less than 0.4\%) were discarded due to processing issues, leaving us with 3490 files.
Due to the fact that the corpus is comprised of more than one source, we  assigned 16 files as the development set, and 16 as the test set, chosen uniformly\footnote{%
Dev files:  [ 217, 435, 651, 868,  1085, 1302, 1519, 1736,
             1953, 2170, 2387, 2604,  2821, 3038, 3255, 3472].
Test files: [ 218, 436, 652, 869,  1086, 1303, 1520, 1737,
             1954, 2171, 2388, 2605,  2822, 3039, 3256, 3473].
             }.
The rest of the files were used as the full training set.
Subsets of this training set were each chosen uniformly too,  
to emulate availability of smaller training sets.
This split departs from \newcite{hong-2018-learn-event-rep}, which used the last 0.4\% (14 files) as the test set, the immediately preceding 0.4\% as the dev set, and the rest for training.

\subsection{Additional test sets}\label{sec:thematic-fit-tests}
We also tested our models on the above-mentioned thematic fit test sets, \textit{without} optimizing on them:

\newterm{Padó-all:} A human-rated thematic fit score dataset collected with psycholinguistic motivations, created by \newcite{pado2007integration} and containing 414 verb-noun-role triplets, where every two triplets differ only in the role, one of \{\term{Arg0, Arg1, Arg2}\}.

\newterm{McRae-all:} A similar dataset with human scores, created by \newcite{mcrae1998modeling}, containing 1,444 such triplets, grouped in pairs similarly, but the roles are only \{\term{Arg0, Arg1}\}, and the words are less frequent  (a harder task).

\section{Experiments}\label{sec:experiments}

\subsection{Baseline Model Configuration}
For our baseline (\vi-based models), we used a multi-task residual network (ResNet) model \citep{he2016deep,jegou2010product}. Our implementation is similar to the best reported model in \newcite{hong-2018-learn-event-rep}, called ResRofa-MT, for ease of comparison.\footnote{We thank Xudong Hong for his help.}
One task was \newterm{role prediction}, given a verb, a (typically noun) word, and zero or more  $\langle$role,word$\rangle$ pairs. 
For example, given `cut' and `knife', predict \textit{Instrument}, or in PropBank's roles: \textit{A3-MNR}.
A second task was \newterm{word prediction} (slot / role filling), given the word's role, the verb,  and the zero or more $\langle$role,word$\rangle$ pairs. 
For example,  given `cut' and \textit{A3-MNR}, predict `knife'.
For both tasks we could optionally provide more input, say, $\langle$\textit{A1},`cake'$\rangle$.
Aside from having different prediction layer per task, the tasks shared the same neural network (and parameters).
Apart from software engineering differences, 
the most notable difference in our implementation is having two separate labels for \textit{missing role} and \textit{unknown role} instead of one for both. The former is used to mark the absence of a certain role from the annotated frame instance. The latter is used as a catch-all for sparser roles not explicitly represented. The baseline role set was comprised of \textit{PRD, Arg0, Arg1, ArgM-TMP, ArgM-LOC, ArgM-MNR}: the predicate, PropBank's arguments 0 and 1,  the temporal, location, and manner modifiers (and \textit{missing role} and \textit{unknown role}).

For faster training time, we used a batch size of 
1024
samples (unless otherwise specified) with the 
risk of too coarse updates. We kept a simple setting of 0.1 learning rate and no decay. 
We applied the same vocabulary pruning to the top 50k most frequent lemma forms as \newcite{hong-2018-learn-event-rep} did.

\begin{table*}[!hbt]
    \centering
    \begin{small}
    \begin{tabular}{cccccccc}
        \hline
        Training & & & & \multicolumn{2}{c}{$\rho_\text{Padó}$}  & \multicolumn{2}{c}{$\rho_\text{McRae}$} \\
        sample & & & & & & & \\
        (\# trials) & Version & Role acc. & Word acc. & final & max & final & max\\
        \hline
        \multirow{2}{*}{0.1\% (3)} & \vi & .8857 \pmx .0009 & .0435 \pmx .0001 & .2760 \pmx .0331 & .2760 \pmx .0331 & .1924 \pmx .0110 & .1968 \pmx .0124 \\
        & \vii & .9102 \pmx .0063 & .1029 \pmx .0007 & .3149 \pmx .0308  & .3257 \pmx .0412  & .1934 \pmx .0044 & .2065 \pmx .0057  \\
        \hline
        \multirow{2}{*}{1\% (5)} & \vi & .9332 \pmx .0006 & .0819 \pmx .0002 & .5150 \pmx .0299 &  .5230 \pmx .0141 & .3142 \pmx .0079 & .3157 \pmx .0069 \\
        & \vii & .9656 \pmx .0001 & .1416 \pmx .0002 & .4850 \pmx .0135 & .4975 \pmx .0141  & .3368 \pmx .0130 & .3398 \pmx .0118\\
        \hline
\multirow{2}{*}{10\% (3)} & \vi & .9419 \pmx .0017 & .0941 \pmx .0005 & .5166 \pmx .0345 & .5368 \pmx .0020 & .3996 \pmx .0206 & .4126 \pmx .0091\\
        & \vii & .9715 \pmx .0010 & .1541 \pmx .0045 & .5229 \pmx .0227 & .5623 \pmx .0227 & .3935 \pmx .0192 & .3981 \pmx .0223 \\
        \hline
\multirow{2}{*}{20\% (3)} & \vi & .9445 \pmx .0003 & .0982 \pmx .0011 & .5219 \pmx .0069 & .5306 \pmx .0073 & .4314 \pmx .0123 & .4381 \pmx .0032 \\
        & \vii & .9733 \pmx .0004 & .1621 \pmx .0048 & .5363 \pmx .0035  & .5494 \pmx .0111 & .4322 \pmx .0232 & .4385 \pmx .0257 \\
        \hline

    \end{tabular}
    \end{small}
    \caption{\vi\  vs \vii: model MTRFv4Res, train \% out of 3490 text files, dev/test size = 16 text files each, 
    batch=1024 unless specified. (\# runs in parentheses).
    \shwTrn{\#itr: number of epochs till early stop.}
    Role/word prediction accuracy (acc.); Spearman's rank correlation ($\rho$) for Padó-all and McRae-all.
    final:score of last saved model; max: maximal score in any epoch.
    }
    \label{tab:v1-vs-v2}
\end{table*}

\subsection{\vi\  vs \vii\ experiments}\label{sec:v1-vs-v2}
We trained and evaluated models in the above configuration  with increasing training set size
 (see Section~\ref{sec:split}), measured in percentage of total number of available training sentences.
For each training set size, we trained few models on \vi\  and same number of models on \vii.

Table~\ref{tab:v1-vs-v2} shows that role prediction accuracy increases with training set size as expected and surpasses 90\% at 1\% of the training set for \vi\ and already at 0.1\% for \vii. At 10\% of the training, it reaches mid-90s for \vi\ and high-90s for \vii, with small further gains at 20\%.%
\footnote{The full dataset size was prohibitive for training given our computing resources.}
The advantage of \vii\ was kept throughout but decreased from over 5-6\% (absolute) at 0.1\% size to 3-4\% at higher training set sizes.
Word prediction accuracy followed similar trends, but \vii--\vi\ gaps there remained about 6\% in all set sizes.%
\footnote{Preliminary experiments, having varying sized of dev and test sets, showed the same pattern: \vii\ advantage.}

We also tested the models on two thematic fit tasks on which the models were not trained, using the additional test sets described in Section~\ref{sec:thematic-fit-tests}. 
Variance over multiple training runs per model was not reported in the relevant literature, so the high variance we see on both  
Padó-all%
\footnote{In Padó-all 
evaluations, test set target \term{Arg2} was mapped to \textit{unknown-role}, since it was not in the model's role set
(McRae-all only has \term{Arg0,Arg1} targets).}
and McRae-all
is a novel finding.%
\footnote{We trained 5 models for each version in small training sizes, and 3 for each in large sizes, due to resource limitations.}
We report both the best- and last-epoch results for these tasks on the rightmost two columns of Table~\ref{tab:v1-vs-v2}.
We see  \vii\ advantage on Padó-max 10\% and 20\% but not at lower sizes. Padó-final results are mixed, particularly at 10\%-training, as are the results on McRae (both last and best).
We see clear training size effects on McRae(best+last), but not on Padó (except in the smallest size).

The \vi\  results on the 10\% and 20\% subsets are our closest replication of \newcite{hong-2018-learn-event-rep}, modulo the above-mentioned role representation.

\begin{table*}[!hbt]
    \centering
    \begin{small}
    \begin{tabularx}{\textwidth}{Xp{.61in}p{.0in}p{1.05in}p{0.98in}p{.80in}p{.88in}}
        \hline
        Name & Role set & \shwTrn{\#itr}  
        & Role acc. \shwTrn{tr/} dev/test & Word acc. \shwTrn{tr/} dev/test & $\rho_\text{Padó}$~final/max & $\rho_\text{McRae}$~final/max \\
        \hline
        
        2Args3Mods  & 
            baseline & 
            \shwTrn{18} & 
            \shwTrn{.9790 /} \textbf{.9653 / .9656} & 
            \shwTrn{.1548 /} .1393 / .1414 &
            .4765 /.4840 & .3205 / .3240 \\
        \hline 
        3Args3Mods & 
            +Arg2 & 
            \shwTrn{18} & 
            \shwTrn{.9753 /} {.9595 / .9596} & 
            \shwTrn{.1714 /} .1544 / .1563 &
            \textbf{.5056 / .5150} & .3340 / .3340 \\
        3Args4Mods & 
            +AM-MOD & 
            \shwTrn{19} & 
            \shwTrn{.9758 /} {.9606 / .9609} & 
            \shwTrn{.1780 /} .1631 / .1661 & 
            .4663 / .4928 & .3261 / .3373 \\
        3Args5Mods &
        +AM-ADV & 
        \shwTrn{19} &
        \shwTrn{.9688 /} .9513 / .9516 & 
        \shwTrn{.1805 /} .1665 / .1691 &
        .4838 / .5024 & .3381 / .3407\\
        3Args6Mods & 
        +AM-DIS & 
        \shwTrn{19} &
            \shwTrn{.9679 /} .9503 / .9510 & 
            \shwTrn{.1828 /} .1683 / .1712 &
            .4742 / .4851 & .3357 / .3370 \\
        3Args7Mods & 
        +AM-NEG & 
        \shwTrn{19} &
            \shwTrn{.9681 /} .9506 / .9512 & 
            \shwTrn{.1874 /} .1742 / .1768 &
            .4808 / .4886 & .3357 / .3385 \\        
        all.args+mods &
            all-roles & 
            \shwTrn{19} &
            \shwTrn{.9632 /} .9450 / .9459 & 
            \shwTrn{.1918 /} \textbf{.1783 / .1810} &
            .4833 / .5109 & .3205 / .3209 \\  
        \hline

        3Args3Mods & 
            +Arg2 & 
            \shwTrn{18} & 
            \shwTrn{.9753 /} .9595 / .9596 & 
            \shwTrn{.1714 /} .1544 / .1563 &
            .5056 / .5150 & .3340 / .3340 \\

        4Args3Mods & +Arg3 & 
            \shwTrn{18} & 
            \shwTrn{.9747 /} .9580 / .9585 &  
            \shwTrn{.1732 /} .1557 / .1582 &
            .5007 / .5119 & .3365 / .3394 \\
        5Args3Mods 
        & +Arg4 & 
            \shwTrn{18} &
            \shwTrn{.9738 /} .9574 / .9576 & 
            \shwTrn{.1737 /} .1559 / .1583 &
            .4901 / .5108 & \textbf{.3473 / .3473} \\
        6Args3Mods 
        & +Arg5 & 
            \shwTrn{18} &
            \shwTrn{.9738 /} .9577 / .9579 & 
            \shwTrn{.1734 /} .1560 / .1582 &
            .4925 / .5237 & .3166 / .3182 \\
         \hline 
    \end{tabularx}
    \end{small}
    \caption{Increasing role set granularity (\vii): model MTRFv4Res, train size = 1\% of 3490 text files; dev/test size, batch size and metrics same as in Table~\ref{tab:v1-vs-v2}.
    Top half: adding next role in descending role frequency.
    Lower half: adding only arguments (skipping modifiers).
    In each half, the roleset in each row is a superset of the  previous row.
    }
    \label{tab:add-arg2}
    \vspace{-6pt}
\end{table*}

\begin{table}[!hbt]
    \centering
    \begin{small}
    \begin{tabular}{rl}
        \hline
        Count & Label \\
        \hline
        2,120,947	&	ARG1	\\
        1,234,063	&	PRD	\\
        1,090,751	&	ARG0	\\
        688,268	&	ARG2	\\
        380,294	&	ARGM-TMP	\\
        257,056	&	ARGM-MOD	\\
        227,040	&	ARGM-ADV	\\
        220,502	&	ARGM-MNR	\\
        194,532	&	ARGM-LOC	\\
        95,724	&	ARGM-DIS	\\
        87,036	&	ARGM-NEG	\\
        68,156	&	ARGM-PRP	\\
        39,780	&	ARGM-DIR	\\
        35,938	&	ARGM-ADJ	\\
        31,004	&	ARG3	\\
        27,850	&	ARGM-CAU	\\
        22,092	&	ARG4	\\
        18,254	&	ARGM-EXT	\\
        13,456	&	ARGM-PRD	\\
        9,108	&	ARGM-LVB	\\
        5,540	&	ARGM-GOL	\\
        3,826	&	ARGM-COM	\\
        3,460	&	ARGM-PNC	\\
        1,686	&	ARGM-REC	\\
        12	&	ARG5	\\
\hline
    \end{tabular}
    \end{small}
    \caption{SRL label counts in dev set}
    \label{tab:srl-label-counts}
\end{table}

\begin{table}[!bt]
    \centering
    \begin{tabular}{|l|r|r|}
        \hline
        \textbf{data set} & \textbf{previous (\vi)}& \textbf{this (\vii)}  \\
        \hline
        \textbf{training 10\%} & 16,889,581  &        20,151,313 \\
        \textbf{dev}           &    766,333  &           915,473 \\
        \textbf{test}          &    767,325  &           919,365 \\
        
        \hline
        
    \end{tabular}
    \caption{Number of frame annotations in \vi\ vs. \vii}
    \label{tab:v1-vs-v2-annot}
    \vspace{-16pt}
\end{table}

\subsection{Roleset Granularity (on \vii)}

Model design includes decisions about features and their granularity. Generally, fine-grained features provide sharper, more accurate distributions of underlying phenomena, but their values, especially at the distribution ``tail'', suffer from higher sparsity, which may lead to difficulty in learning them well and hence lower performance overall.
This tension between granularity and sparsity exists also for the case of role set granularity.
Many studies have chosen the coarser side, using only few thematic roles. For example, \newcite{hong-2018-learn-event-rep}, which we take as our main baseline, use only 2 arguments and 3 modifiers, within the PropBank framework---in itself already a framework with  a coarse role set (compared to VerbNet and FrameNet).

We tested the effect of increasing role set granularity. In other words:
is less (less coarse role set) more (higher quality)? 
We analyzed the distribution of the roles in the dev set (see Table~\ref{tab:srl-label-counts}) and expanded the role set one role at a time, according to their frequency (more frequent first). For training run time economy, all models in this subsection were trained on 1\% of the \vii\ data.

The first semantic role to be added was \term{Arg2}. Note that now in Padó-all 
evaluations, \term{Arg2} is no longer mapped to \textit{unknown-role}.
See Table~\ref{tab:add-arg2}.
It turns out that while role set prediction accuracy slightly dropped, word prediction accuracy actually improved by more than 1\%. Same trend held also for Padó-all (about 3\% gain in Spearman's correlation) and McRae-all (over 1\%).
This stands in contrast to preliminary experiments  on \vi.\footnote{Hong, personal communication.}

Next down the role list, we added \term{ArgM-MOD}.
While we saw a slight gain in word prediction accuracy, we saw a drop in Padó-all and perhaps a slight drop in McRae-all. However, adding \term{ArgM-ADV} resulted in a drop in role prediction but gains in the thematic fit tasks.
Adding \term{ArgM-DIS} resulted in some drop on the thematic fit tasks, while adding \term{ArgM-NEG} yielded relative gains in word prediction and Padó-all.
Adding all  roles to the model (\term{all.args+mods}) yielded mixed results: a further small drop in role prediction (in fact, no model outperformed the baseline on this task), a pronounced gain in word prediction (highest result even compared to Table~\ref{tab:v1-vs-v2}!), and lower scores on the thematic fit tasks compared to adding only \term{Arg2}.

Modifiers are often more freely optional, and are not considered core arguments of the semantic frame.
We therefore also tested the effects of only adding core arguments to the baseline roleset (see bottom part of Table~\ref{tab:add-arg2}).
Adding \term{Arg3} resulted in no much change in any task, compared to \term{+Arg2}.
Adding \term{Arg4} yielded a gain on Padó-all last result (but not the maximal result).
Adding \term{Arg5} resulted in a clear drop on the thematic fit tasks, which is expected: \term{Arg5} is very sparse, so its learnability is low, and roleset confusability is higher due to increased number of roles.
However, we take differences on Padó-all and McRae-all with a grain of salt, given the above-mentioned high variability.

\section{Discussion and Analysis}
The advantage of \vii-trained models over \vi-trained models for role and word prediction in every training size 
undermines
the NLP community's widely-held working assumption that mediocre is always preferable 
at scale.
It supports our hypothesis that sometimes
better annotations yield better results, even at scale, 
compared to baselines of reasonable mediocrity.
While we do not know if this holds also in the limit, this finding is worth keeping in mind even with today's very large datasets. 

To validate our claim that \vii\ annotations are much better than \vi,
we randomly sampled 
8
sentences, and counted
the difference in number of frames, 
number of roles/arguments, 
and number of wrong roles between the two datasets. 
\vii\ had a clear advantage over \vi\ in identifying frames and roles, with almost no cases in which \vi\ did better. 
This advantage held in both the number of cases and the number of sentences with offending cases (63--75\%).
Both \vi\ and \vii\ had few wrong roles, similar in number, with perhaps a slight advantage to \vi\ (one case).%
\footnote{Note that for verification speed, we only verified correctness of \vii's \term{Arg0, Arg1}; the rest we assumed correct.}
Due to the small sample size and evaluation method, we take the findings above as mainly qualitative, but still strongly supporting our assumption of \vii\ advantage.

Was \newterm{quality} the only factor?
Several aspects here:
\textbf{(a)} better argument span and role prediction in LSGN in \vii\ compared to SENNA in \vi, together with 
\textbf{(b)} the greater \textit{quantity} of predicted frames with LSGN,
compounded by 
\textbf{(c)} better parsing quality of spaCy in \vii\ compared to Malt in \vi, and 
\textbf{(d)} Morfette's better lemma analysis.

As for \newterm{quantity},  
it turns out
it may have also played a role:
we compared the number of semantic frame annotations in the 10\% training subset, and found out \vi\ has less than 84\% of 
\vii's,
for the same underlying sentences (Table~\ref{tab:v1-vs-v2-annot}). 
We  call it \newterm{frame quantity}  to tease it apart from  \newterm{sentence quantity}: the number of (underlying) sentences used for training. Perhaps the better SRL and parser quality contributed both to the increase in number of frames as well as to the number of 
correctly extracted syntactic head words (one per argument, using better aligned parses).

Was the \vii\ advantage mainly due to frame quantity? 
1\%-training \vii\ outperforming 10\%-training \vi\ on role and word prediction,
even though the former was trained on an 
eighth of the number of frames 
(and tenth of the sentences), suggests otherwise. This trend repeated with
0.1\%
\vii\ outperforming 1\%
\vi\ on word prediction.
Higher quality annotations resulted in large \textit{savings} in training sentence quantity  
for similar prediction quality.

How does our implementation fare compared to \newcite{hong-2018-learn-event-rep}?
Our \vii\ maximal result on Padó-all (59.9\% best single run, 56.2\% averaged) outperforms their reported 53\% with only 10\% of their data.
On McRae-all our maximal result (45.9\% best single run, 43.9\% averaged)
outperforms their reported 42.5\%,
despite having less frequent 
pred-arg
combinations.
Our \vii\ outperforms their reported 94.7\% role 
accuracy, even at 1\% of their training data, presumably
due to our 
separating the 
\term{unknown role} from \term{missing role}. 

Could we have reached even higher results? Our model setting is on the simpler side with only two tasks, one of which is role prediction with accuracy approaching 100\% (therefore, after a few iterations,
largely only the other task  affects the learning). More complex models or additional tasks (and/or modern word embeddings) are likely to do even better on the word prediction and the thematic fit tasks. 
However our focus in this work was not on creating the best model, but on exploring the effects of annotation quality and quantity up to large scale.

Did the effort of creating \vii\ pay off also for the \textbf{psycholinguistic task}? We see a clear, but rather small gain in Padó-all on the larger subsets (10\%, 20\%). In McRae-all the gain is actually on the smaller subsets, and goes away on the larger ones. Therefore, we conclude that 
for
improving indirectly-supervised psycholinguistic tasks, the cost-effectiveness of this exercise is questionable, but it still suggests that progress can be made in resource-constrained environments through limited improvements in label accuracy.

As for \newterm{annotation granularity}, adding \term{Arg2} to the baseline's role set showed clear gains across the board (except perhaps in role prediction), which may seem surprising at first: 
\textbf{(a)} the role prediction task is now harder (larger role set) yet accuracy did not drop by much. This could be due to the increased thematic homogeneity in the catch-all role tag.
\textbf{(b)} Word prediction (slot-filling) was not expected to be affected, since the embeddings are not relearned in our setting. But gains may show if many words tend to assume only certain roles (e.g., be Arg2-centric).
\textbf{(c)} Role prediction in Padó-all should have also been harder, therefore showing lower correlation scores. But recall that baseline scores are not directly comparable here because all roles but Arg0 and Arg1 were mapped to Arg2 for the evaluation of Padó-all, a mapping which was no longer needed once we added Arg2 to the role set.
\textbf{(d)} gain in McRae-all is surprising at first because McRae-all only has Arg0 and Arg1 targets, so adding Arg2 could only distract from these targets. But recall we have a catch-all role, and adding Arg2 to the role set made the catch-all role distribution more focused, and therefore presumably less prone to mix-ups with Arg0 and Arg1 -- especially since McRae-all contains less frequent words, which makes them harder to learn.

Adding the next modifiers in order of descending frequency (top part of Table~\ref{tab:add-arg2}) yielded 
consistent monotonous gains in word prediction, with the \term{all-roles} model, trained on 1\% of the training set, performing even better than larger subset models in Table~\ref{tab:v1-vs-v2}. This seems to support  finer-grained representation. However, curiously, adding only core arguments (lower half of Table~\ref{tab:add-arg2}) did not make a noticeable difference on this task.
Adding the next modifiers after \term{Arg2} did not improve Padó-all, which is expected, since it only contains arguments 0-2.
But the top result of \term{+Arg4} on McRae-all is surprising: this test only has \term{Arg0, Arg1} targets. We suspect this result is an outlier even given the high variance, but further investigation is in order.

\paragraph{Ethical considerations}
This work used two large corpora (the 
BNC and ukWaC);
hence 
 it is not practical to completely account for all the data in the corpus. 
The BNC is a curated corpus, but 
part of their transcribed conversations were recorded without prior consent of all recorded individuals. This is no longer an acceptable conduct in Great Britain and many other countries.
Our annotated corpus (\vii) clearly marks the source of each sentence, so those who wish to exclude BNC data can easily do so.
Insofar as future work keeps that in mind, we believe there to be minimal scope for direct misuse of our results.

\section{Conclusions and Future Work}
We set out to test  the NLP community's widely-held assumption that mediocre linguistic annotation at scale is as good as better annotation.
We saw that models trained on better lemmas, syntactic parses, and SRL tags (our \vii) did better than the baseline (\vi) using older technology at all training set sizes, and even at scale -- on both (directly supervised) role and word prediction.
We also saw ``training dataset savings'' potential: training on smaller sets with better annotations yielded sometimes better results than training on datasets with less advanced annotations that were~2 to~10 times larger in size (with McRae-all being a notable exception).
To better understand that, we teased apart contributions of annotation quality and quantity, and their interplay. We further teased apart sentence quantity from frame quantity.

We saw a 
high variance in thematic fit estimation,  to the point where in one task (Padó-all) \vii\ advantages over \vi\ only showed in larger training set sizes, while in another task (McRae-all) no advantages were seen. Given the small-to-no gains in these tasks relative to \vi, the cost-effectiveness of re-annotating with better tools is questionable for the indirect supervision setting.

We saw  all tasks benefited from increasing the training sentence set size, at least until 10\% of our large corpus (except perhaps Padó-all beyond 1\%). Future work should check if larger sizes can yield even better results.
Even at 10\% of the training, our \vii\ model set a new record in indirectly supervised thematic fit estimation on Padó-all,
and at 20\% a new record also on McRae-all (both \vi, \vii).

We also saw that refining the semantic role set granularity helps in thematic fit tasks (and word prediction). On Padó-all, best results  were achieved already by adding \term{Arg2}, but surprisingly  on McRae-all 
by adding \term{Arg2-4}. Adding all roles yielded best results on word prediction.
These are novel results.

Last, but not least, we introduced a new open-data annotated corpus.%
\footnote{Available 
at 
\url{http://yuvalmarton.com/RW-Eng}}
We believe this new corpus will be useful to the NLP community beyond our reported experiments. Future work involves studies with existing and new annotation layers,
e.g., combining \vi+\vii\ parses, \vi+\vii\ SRL tags, and replicating these experiments with various  word embeddings and different network architectures.
Future work should also explore different optimization objectives or additional tasks in the multi-task setting,
since role prediction seems too easy (reaches accuracy in high 90s early on), while the word prediction objective may be too hard,
although the latter can be ameliorated with better word embeddings and a loss function based on the vector distance between predicted and target words. 
We also plan to add new thematic fit tasks with multiple simultaneous role-fillers  \citep{bicknell2010effects,vassallo2018event}.

\section{Acknowledgements}
A. Sayeed's involvement in this research was funded by a Swedish Research Council (VR) grant (2014-39) for the Centre for Linguistic Theory and Studies in Probability (CLASP).
We thank the reviewers for helping us make this paper clearer.

\section{Bibliographical References}\label{reference}

\bibliographystyle{lrec2022-bib}
\bibliography{thematic-fit,mtrf,custom}

\end{document}